\documentclass[sn-mathphys,Numbered]{sn-jnl}

\usepackage{graphicx}%
\usepackage{multirow}%
\usepackage{amsmath,amssymb,amsfonts}%
\usepackage{amsthm}%
\usepackage{mathrsfs}%
\usepackage[title]{appendix}%
\usepackage{xcolor}%
\usepackage{textcomp}%
\usepackage{manyfoot}%
\usepackage{booktabs}%
\usepackage{algorithm}%
\usepackage{subcaption}%
\usepackage{algorithmicx}%
\usepackage{algpseudocode}%
\usepackage{listings}%
\usepackage{amssymb}
\usepackage{latexsym}
\usepackage{url}
\usepackage{amsmath,amssymb,amsfonts,bm}
\usepackage{booktabs}
\usepackage{arydshln}
\usepackage{tabulary}
\usepackage{graphicx,epstopdf}
\usepackage{textcomp}
\usepackage{subcaption}
\usepackage{booktabs}
\newcommand{\revision}[1]{\textcolor{black}{#1}} 

\begin{document}

\title{Robotic CBCT Meets Robotic Ultrasound}

%

\author{\fnm{Feng} 
\sur{Li$^1$}}
\email{feng.li@tum.de}

\author{\fnm{Yuan} 
\sur{Bi$^1$}}
\email{yuan.bi@tum.de}

\author{\fnm{Dianye} 
\sur{Huang}}
\email{dianye.huang@tum.de}

\author*{\fnm{Zhongliang} 
\sur{Jiang}}
\email{zl.jiang@tum.de}

\author{\fnm{Nassir} 
\sur{Navab}}
\email{nassir.navab@tum.de}




\affil{CAMP, Technical University of Munich, Munich, Germany}
\affil{Munich Center of Machine Learning, Munich, Germany}
\affil[1]{Equal Contribution}
\abstract{

\textbf{Purpose:}
The multi-modality imaging system offers optimal fused images for safe and precise interventions in modern clinical practices, such as \revision{computed tomography - ultrasound (CT-US)} guidance for needle insertion. However, the limited dexterity and mobility of current imaging devices hinder their integration into standardized workflows and the advancement toward fully autonomous intervention systems. In this paper, we present a novel clinical setup where robotic \revision{cone beam computed tomography (CBCT)} and robotic US are pre-calibrated and dynamically co-registered, enabling new clinical applications. This setup allows registration-free rigid registration, facilitating multi-modal guided procedures in the absence of tissue deformation.

\textbf{Methods:}
First, a one-time pre-calibration is performed between the systems. To ensure a safe insertion path by highlighting critical vasculature on the 3D CBCT, SAM2 segments vessels from B-mode images, using the Doppler signal as an autonomously generated prompt. Based on the registration, the Doppler image or segmented vessel masks are then mapped onto the CBCT, creating an optimally fused image with comprehensive detail. To validate the system, we used a specially designed phantom, featuring lesions covered by ribs and multiple vessels with simulated moving flow.

\textbf{Results:}
The mapping error between US and CBCT resulted in an average deviation of $1.72\pm0.62$ mm. A user study demonstrated the effectiveness of CBCT-US fusion for needle insertion guidance, showing significant improvements in time efficiency, accuracy, and success rate. Needle intervention performance improved by approximately 50\% compared to the conventional US-guided workflow.

\textbf{Conclusion:}
We present the first robotic dual-modality imaging system designed to guide clinical applications. The results show significant performance improvements compared to traditional manual interventions.
}

\maketitle

\section{Introduction}
\par
Traditional medical interventions often rely on a single-image modality such as ultrasound (US), X-ray, or magnetic resonance (MRI). These single modalities have their own advantages and favorable use cases in reality. However, it is also commonly known that they often suffer from limitations such as sub-optimal contrast, limited image view, sensitivity to tissue properties, and also real-time performance. To tackle this problem, the dual-modality or multimodal imaging fused imaging system has been seen as a promising solution for further enhancing the safety, precision, and stability of future image-guided interventions. Pioneering examples are interventional single photon emissions tomographie-computed tomography (SPECT-CT) imaging system~\cite{gardiazabal2014towards} and CT-US system~\cite{monfardini2018ultrasound}. Such dual-modality systems can provide complementary information in terms of field of view or optimal visualizations of different tissues to have an ideal fused image. For example, the CT-US or cone beam computed tomography-US (CBCT-US) fusion is desired for guiding and tracking the needle insertion procedure, where the CBCT provides high-quality 3D images of hard tissues while US excels in soft tissue visualization and provides real-time, radiation-free guidance. This fused modality allows clinicians to leverage both static pre-operative 3D views and dynamic intraoperative imaging. The combination of US and CBCT has proven effective in applications such as liver ablation~\cite{monfardini2018ultrasound}, renal ablation~\cite{monfardini2021real}, and radiation therapy for uterine cervix cancer~\cite{mason2019combined}.

\par
However, deploying two imaging devices in the operation room has practical challenges over the limited working space, altering the standard intervention workflow. To tackle this problem, state-of-the-art (SOTA) is considering motorizing image devices such as mobile robotic CTs~\cite{kim2019design, weir2015dosimetric, ebinger2015mobile} and robotic C-Arms~\cite{tsang2015real,tanaka2024low}. Recently, a mobile cone-beam computed tomography (CBCT) scanner called LoopX (medPhoton, Austria) was designed with great dexterity in both translational and rotational directions to meet the requirement of mobility~\cite{karius2024first}. By automating workflows, Loop-X is advancing mobile imaging with functionality that will be desirable for intervention procedures, particularly in combination with intraoperative US imaging modality. 

\par
To further eliminate the down facts of traditional free-hand US scanning in terms of reproducibility and standardization, the US probe is desired to be maneuvered by a robotic arm, particularly when we need to position the probe precisely to guide the needle insertion process to a target, such as the tumor in ablation procedure~\cite{li2024invisible, jiang2024needle}. Recent advances in the field of robotic US have demonstrated the superior performance of robots over humans in terms of precision, stability, and reproducibility~\cite{jiang2023robotic, bi2024machine, von2021medical, huang2024robot_qin, li12autonomous}. The integration of robotic CBCT and robotic US systems holds significant potential for future surgical applications, while this combination has yet to be explored in the research community.  

\par
In this work, we present the first robotic CBCT-US dual-modality imaging system, where the robotic CBCT and robotic US images are pre-calibrated and dynamically co-registered, which enables novel clinical applications in which a registration-free rigid registration allows for multi-modal guided procedures in the absence of deformation. Unlike previous approaches, our system leverages robotic precision to automatically determine and maintain the optimal probe positioning, ensuring accurate needle insertion without the need for manual adjustments. This capability is particularly valuable in procedures like liver ablation, where precise targeting of targets and avoiding critical structures such as blood vessels are crucial. To prove the fused image with ample information, we also map the Doppler images into the 3D CBCT volume or slice based on the registration. The effectiveness of the proposed system was validated on a carefully designed phantom with lesions covered by ribs and multiple vessels, including mimicked moving flow. \revision{Although deformation was not considered, the system primarily focuses on offering a robust initial registration framework that can be further optimized for future research on deformable registration methods.}

\section{Methods}
\begin{figure}[h]
\centering
\includegraphics[width=0.98\textwidth]{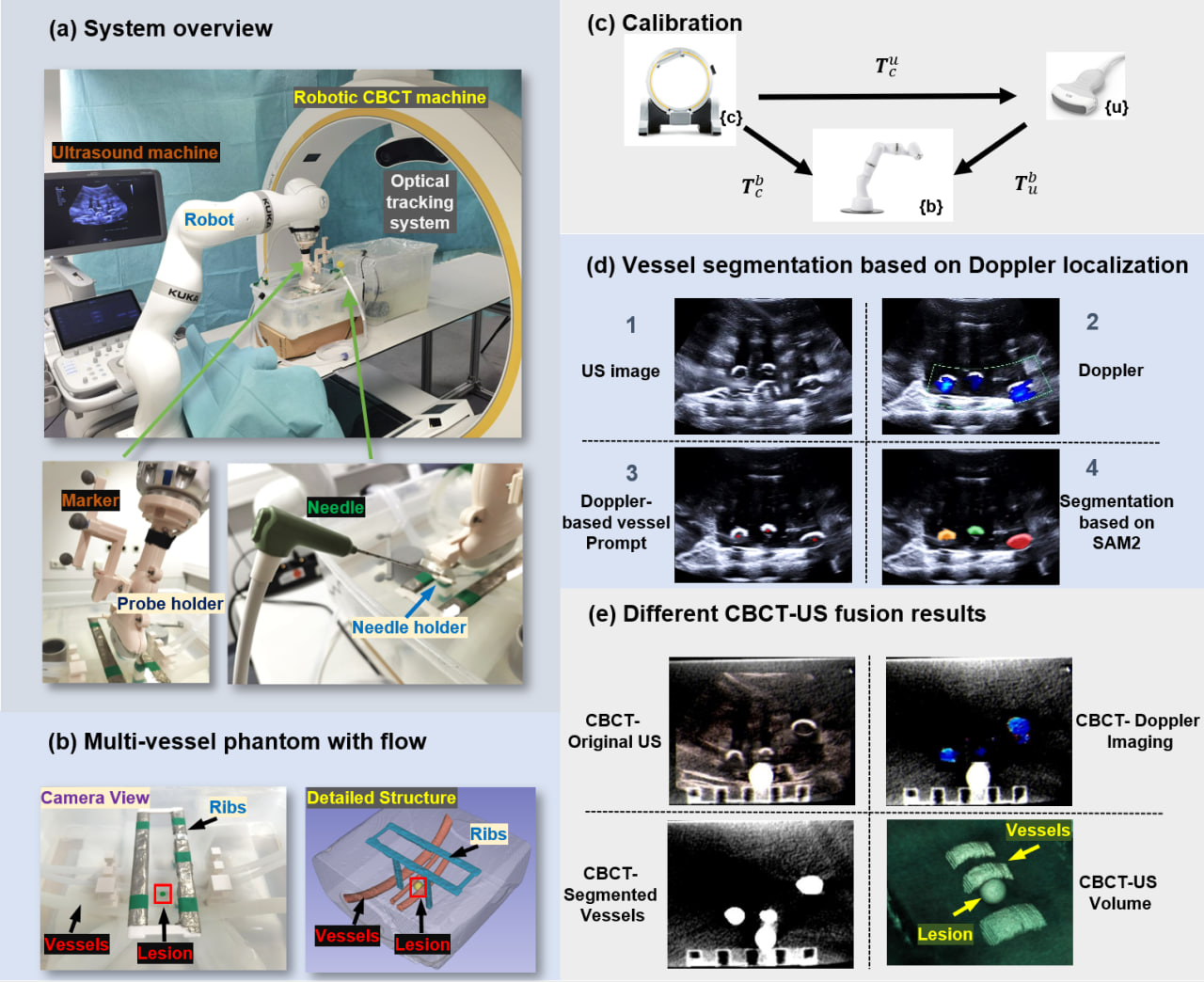}
\caption{Overview of the robotic CBCT-US system and fusion results.}\label{fig:overview}
\end{figure}

An overview of the proposed robotic CBCT-US system is shown in Fig.~\ref{fig:overview}(a). It consists of two main parts, a robotic US system and a robotic CBCT system. The robotic US system is composed of a KUKA arm (LBR iiwa 14 R820, KUKA GmbH, Germany) and a Siemens US machine (ACUSON Juniper, Siemens Healthineers, Germany), where the US probe (5C1, Siemens Healthineers, Germany) is rigidly attached to the end-effector of the robot using 3D printed probe holder. The US images are captured using a frame grabber (Epiphan Video, Canada), so that the images are accessible from the controlling computer. \revision{Each frame has a resolution of $880 \times 660$. During the whole experiment, US acquisition parameters are fixed.} The robotic CBCT system utilizes a commercial mobile ``ImagingRing'' (LoopX, medPhoton, Austria), with six degree of freedoms (DoFs), three for translational and rotational movements of the device on the ground, two for rotations of the X-ray source and detector, and one for rotation of the ``ImagingRing''. Additionally, LoopX is equipped with an integrated optical tracking camera mounted on the upper part of the ``ImagingRing'', as shown in Fig.~\ref{fig:overview}(a). In order to co-register the two robotic systems, a hand-eye calibration is performed between the robot arm and the optical tracking camera of LoopX. Once calibrated, the two systems are able to communicate and interact with each other.

\par
\revision{The initial setup we showed in this work can be treated as a foundational system proposal for the future robotized intelligent imaging system}, where multi-modality imaging technologies can work together, complementing each other to enhance the precision and flexibility of medical procedures. With such a system, the radiologists would no longer be limited to viewing a static image but could perform real-time acquisition and measurements in the areas of interest, e.g., Doppler imaging, which is mapped on top of the CT/CBCT image. This approach has the potential to streamline the current surgical workflow.
We simulated a challenging needle insertion scenario for system validation. The current surgical workflow of abdominal ablation often involves both CT and US, where CT is primarily used for pre-operative planning and occasionally for intra-operative guidance through CT fluoroscopy, while US is mostly utilized as real-time continuous imaging feedback during needle insertion. In the proposed robotized setup, the entire needle insertion process can be fully or semi-automated upon surgeons' demands, largely reducing radiation exposure while enhancing the efficiency and accuracy of needle placement. This automation streamlines the workflow and minimizes the need for continuous manual adjustments during the procedure.

\par
A phantom that simulates the complex anatomical structures of liver is built as shown in Fig.~\ref{fig:overview}(b). Three tubes with inner diameters of 7mm, 8mm, and 16mm, respectively, with a wall thickness of 1mm, \revision{reflecting typical abdominal blood vessel sizes, such as the renal vein (5–8 mm), hepatic vein (6–10 mm), and abdominal aorta (15–25 mm)}, are placed inside the phantom with a water pump circulating water through them to simulate vasculatures. Below the three tubes, a glass ball with a diameter of 10mm is positioned to represent a lesion inside the liver, while two 3D-printed long strip parts are placed near the surface to mimic the ribs. For simplification, all the mimicked ribs are put into a water tank. The procedure begins by acquiring a CBCT scan of the target volume, in this case, the phantom. Then the lesion can be located inside the volume. By applying the registered transformation between the CBCT volume captured by LoopX and the robot base, the robotic US system can localize both the lesion and the positions of the ribs. The robot is then controlled to place the US probe to a proper position that can visualize the lesion while avoiding any occlusions caused by ribs. Next, in order to localize the distribution of vessels near the lesion, the probe is controlled to perform a fan motion to reconstruct the vasculatures with the help of Doppler imaging. The reconstructed vessels are then mapped to the CBCT volume to create a multi-modality imaging volume. 

\par
In the semi-automated setup, the surgeons can define the needle insertion trajectory based on the fused CBCT-US volume, ensuring it avoids any intersections with vessels. In contrast, the fully automated approach can directly provide an optimal needle insertion trajectory to the center of the lesion while maximizing the distance from the vessels. Once the trajectory is defined, the robotic arm positions the US probe to the correct pose where the needle holder attached besides the US probe can guide the needle insertion to follow the planned trajectory. 

\subsection{System Calibration}\label{sec:calibration}


The registration between the robotic CBCT and the robotic US is performed via a hand-eye calibration process. An optical marker is rigidly attached to the 3D-printed US probe holder. The final goal is to determine the transformation between US imaging coordinate system $\{u\}$ and CBCT coordinate system $\{c\}$ as depicted in Fig.~\ref{fig:overview}(c).

\par
Let $ T^{e}_{b} $ denote the transformation from the robot base $\{b\}$ to the robot end-effector $\{e\}$, which can be determined by forward kinematics of the robotic arm, and let $ T^{o}_{m} $ denote the transformation from the optical marker $\{m\}$ to the optical tracking camera of LoopX $\{o\}$, which can be given by the camera directly. Then the hand-eye calibration process can be formulated as follows:

\begin{equation}
\begin{split}
T^{e(1)}_{b} \, T^{b}_{o} \, T^{o(1)}_{m} & = T^{e(2)}_{b} \, T^{b}_{o} \, T^{o(2)}_{m} \\
(T^{e(2)}_{b})^{-1} \, T^{e(1)}_{b} \, T^{b}_{o} & = T^{b}_{o} \, T^{o(2)}_{m} \, (T^{o(1)}_{m})^{-1} \\
AX & = XB
\end{split}
\label{eq1}
\end{equation}
where $ T^{b}_{o} $ represents the transformation between the robot base coordinate system and the camera coordinate system of LoopX. The transformation $ T^{o}_{c} $ is determined by a camera-CBCT registration as described in~\cite{karius2024first}, while US calibration is also performed following~\cite{jiang2021autonomous} to specify $ T^{u}_{b} $, which depicts the transformation between US imaging coordinate and robot end-effector. Then transformation $ T^{u}_{c} $ can be expressed as :


\begin{equation}
\begin{split}
T^{u}_{c} & =  T^{u}_{b} \, T^{b}_{o} \, T^{o}_{c}
\end{split}
\label{eq2}
\end{equation}

\par
\revision{An automatic hand-eye calibration system was developed for quick and efficient setup. The process includes three steps: defining the robotic end-effector's movement range, automatically sampling paired poses, and solving the calibration matrix using the "Tsai-Lenz" method ~\cite{tsai1989new}. First, 4-6 poses along the workspace borders are manually sampled to define the range. Then, the end-effector moves randomly within this range, recording 30 paired poses for calibration.}
\par
Notably, this registration can be maintained even if the LoopX system moves, as the integrated tracking sensors continuously monitor its movements. The corresponding registration matrix can be updated by applying the transformation measured by the LoopX system, ensuring that the alignment between the CBCT and US imaging coordinate systems remains accurate despite any repositioning of the LoopX during the procedure. This capability ensures consistent imaging fusion and precise navigation, even in dynamic environments.

\subsection{Vasculature Mapping Based on Doppler Imaging}
Doppler imaging is a commonly utilized US-specific imaging technique in clinical practice for vessel localization and measurements~\cite{jiang2023dopus}. In this section we present a simple framework that is able to automatically localize and map the extracted vasculatures in US to the CBCT image utilizing Doppler images as guiding signal. The Doppler signal is treated as a strong indicator for the presence of vessels. As shown in Fig.~\ref{fig:overview}(d) The vessel localization process begins with the extraction of the colored regions from color Doppler images. 
\revision{To minimize noise in the water tank, only the Doppler signals larger than 10 mm$^2$ are kept, ensuring compatibility with the blood vessel dimensions in the targeted application scenario}. The centerpoint of each extracted components are then served as the prompt for the Segment Anything Model 2 (SAM2) to initiate segmentation process. 
Then when the robot is performing a fan-motion, the SAM2 continuously segments the vessels and reconstruct it in 3D.
\revision{Fine-tuning of SAM2 was not conducted, as the vessel borders in the water phantom were clearly visualized in US.}



\par
After scanning and segmentation, each US frame can be mapped to the CBCT volume using the known transformation $ T^{u}_{c} $ from US to CBCT. Based on the position of the segmentation mask in the US images, the vascular pixels are transferred into corresponding voxels within the CBCT volume. Once all pixel-to-voxel mappings are completed, the vascular structure is reconstructed within the CBCT volume, resulting in an enhanced CBCT volume as shown in Fig.~\ref{fig:overview}(e) (bottom right). As shown in Figure 2, the US image, CBCT-US fusion slice, and the enhanced CBCT volume can be shown in our visualized robotic interactive system in real time. It is evident that all critical anatomical structures relevant to liver ablation are clearly visualized, providing a solid foundation for precise needle insertion planning. This detailed presentation has the potential to enhance the accuracy and safety of the procedure by ensuring key structures, such as blood vessels and the lesion, are accurately identified and considered in the planning process.

\par
Apart from mapping the segmented vessel masks to the CBCT volume, it is also possible to directly map the original US images to the CBCT as shown in Fig.~\ref{fig:overview}(e) (top left). Furthermore, the Doppler image, containing the flow information, can also be fused with the acquired CBCT, allowing the surgeons to visualize blood flow patterns alongside anatomical structures, as shown in Fig.~\ref{fig:overview}(e) (top right). Based on the enhanced visualization results, anatomical information from both modalities are well conserved, providing a comprehensive view for more accurate surgical planning.


\subsection{Needle Trajectory Planning}\label{sec:needle_planning}
With the fused visualization of CBCT and US, we also present a simple yet effective automatic needle trajectory planning pipeline. The whole planning process is divided into two steps: out-of-plane localization and in-plane localization. The plane is referred as the US imaging plane. Since the needle is inserted through a needle holder which is rigidly attached to the US probe, therefore, the correct placement of needle can be to finding the accurate pose of the US probe. 

\par
In the out-of-plane localization phase, the axial slice in the CBCT volume that passes through the center of the lesion is automatically allocated based on the lesion segmentation results from CBCT. Then at the in-plane localization phase, based on the calibration matrix between LoopX and robotic US, the US probe is controlled to overlap the US imaging plane with the selected CBCT slice while the lesion is placed in the middle of the US image, ensuring the probe to be orthogonal to the scanning surface, which in the presented setup is the water surface.
Then in the in-plane localization phase, utilizing the calibration matrix between the LoopX system and the robotic US, the US probe is controlled to align the US imaging plane with the selected CBCT slice. Meanwhile the lesion is centered in the middle of the US image. Additionally, the probe is positioned to be orthogonal to the scanning surface, which in the presented setup is the water surface. The needle planning problem is then simplified to an optimization problem of finding a line ($y=kx+b$), which represents the needle insertion trajectory on a 2D plane. The objective is to maximize the distances from this line to the center of each vessel while satisfying specific constraints. These constraints ensure that the planned trajectory passes through the center of the lesion, and the in-plane rotation angle of the US probe is limited to $\pm15^{\circ}$, 

\begin{equation}
\begin{array}{rll}
\mathop{max} \limits_{k,b}~~~~  & \sum_{i=1}^{N} \dfrac{|k \cdot x_{i} - y_{i} + b|}{\sqrt{1 + k^{2}}} &, (x_i,y_i)\in \{\text{vessel centers}\} \\
\text{subject to}~~~~ & k \cdot x_{l} + b - y_{l} = 0 &, (x_l,y_l)\in \{\text{center of lesion}\}\\
& k_{\text{max}} \geq k \geq k_{\text{min}}
\label{eq:optimization}
\end{array}
\end{equation}
where $N$ is the total number of the vessel in the 2D US slice.
\revision{The probe rotation was limited to $\pm15^{\circ}$ to prevent the needle from becoming nearly parallel to the surface, due to the fixed insertion angle of the needle holder, and to maintain proper contact between the convex probe and the scanning surface.} After solving this optimization problem with sequential quadratic programming, the robot is controlled to move the probe to align the insertion trajectory of the needle holder with the planned insertion trajectory.

\revision{
The needle holder can secure a needle at an insertion angle of approximately $39^{\circ}$, as shown in Fig. 2(a), with the corresponding US image displayed in Fig. 2(b). Regardless of probe movement, the needle's position remains fixed in the captured US image. Once the system predicts the insertion angle, the robotic arm automatically adjusts to the appropriate position, ensuring that the needle reaches the target lesion, as shown in Fig. 2(c) and Fig. 2(d).
}

\begin{figure}[h]
\centering
\includegraphics[width=1\textwidth]{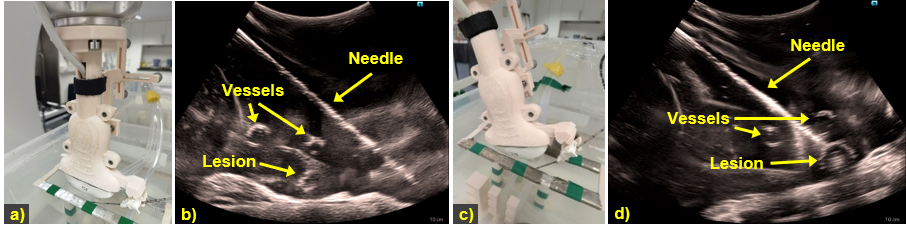}
\caption{\revision{(a) US probe positioned perpendicular to the phantom, showing the initial needle setup. (b) The US image showing the needle inserted into the phantom with the US probe perpendicular to the phantom. (c) Probe and needle positions after the robotic arm's movement. (d) US image after movement, with the needle indicating the predicted path toward the target lesion.}}\label{fig:sketch}
\end{figure}

\par
The designed pipeline is a tailored strategy specifically developed for this particular use case. More generalized, 3D-based trajectory planning could also be implemented to handle a wider range of scenarios. However, the primary goal here is to demonstrate the precision of the proposed robotic CBCT-US system in a straightforward automatic needle placement task, highlighting its potential for accurate interventions.

\section{Results}\label{sec3}
\begin{figure}[h]
\centering
\includegraphics[width=1\textwidth]{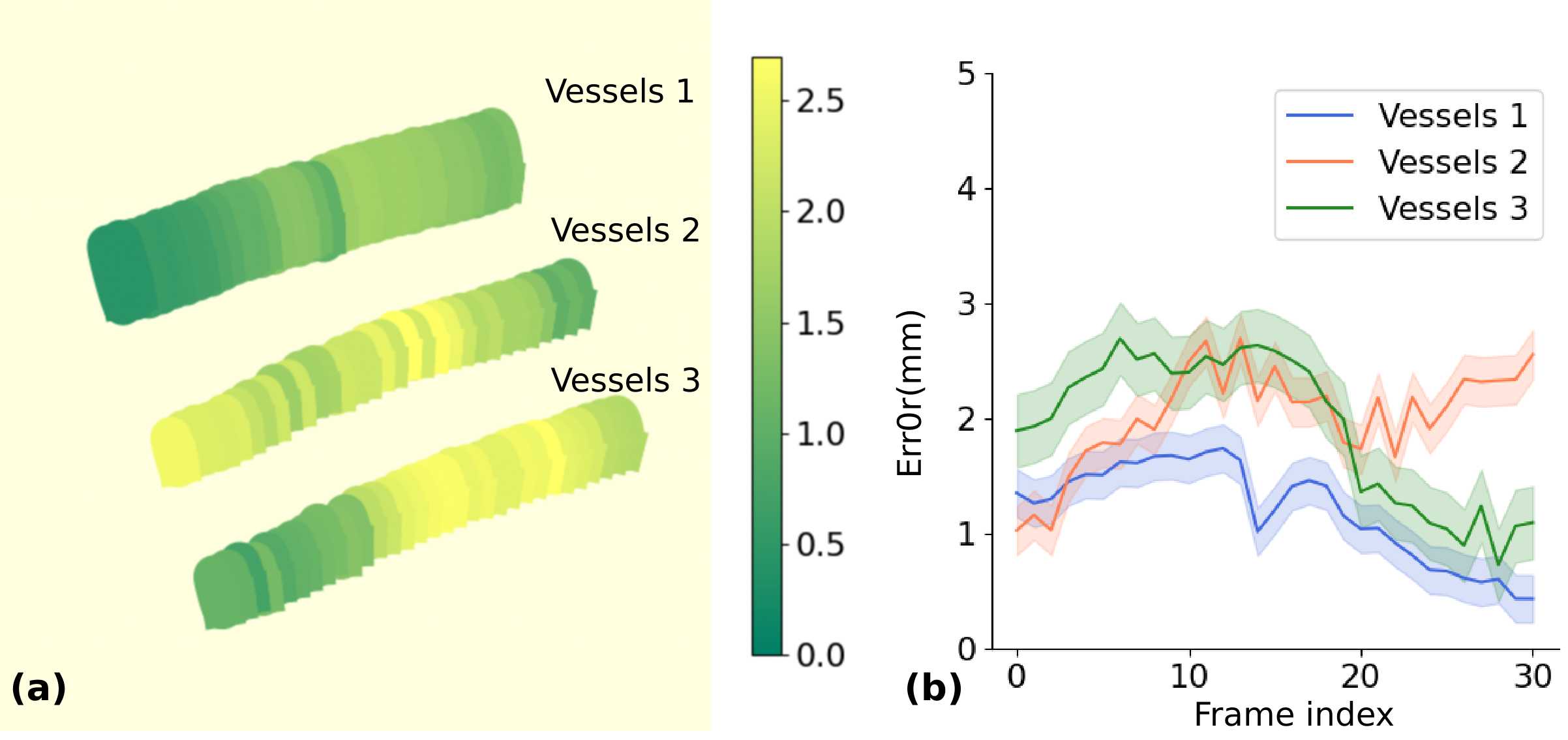}
\caption{(a) Mapping error of blood vessels shown in color gradient. Green means the error is low, while yellow represents the error is high. (b) Mapping errors over frames.}\label{fig:mapping_error}
\end{figure}

\subsection{CBCT-US mapping performance}\label{subsec2}
The mapping accuracy between the US and CBCT modalities is evaluated by measuring the matching errors of the three vessels \revision{, and the robot-US calibration error is also implicitly assessed through the mapping error, assuming the precision of the robot-US calibration has been ensured}. The vessels are first segmented from the tracked US images, and the centerlines of the vessels are extracted. These extracted centerlines are then mapped to the CBCT coordinate system using the calibration matrix. Since the mimicked vessels are made from rubber, they are clearly visible within the CBCT volume, which enables the direct localization of the vessel positions in CBCT. The mapping error is determined by calculating the distance between the vessel centerlines extracted from US and those extracted from CBCT, providing an assessment of the alignment accuracy between the two modalities. As shown in Fig.~\ref{fig:mapping_error}(a), the color gradient on the vessels represents the mapping error between the US-extracted centerlines and the corresponding vessel centerlines extracted from the CBCT volume. The error is visualized along the length of each vessel, with lighter colors indicating higher errors and darker shades indicating lower errors. It can be observed that the error distribution is relatively consistent along each vessel. Fig.~\ref{fig:mapping_error}(b) shows the mapping error over multiple frames for each of the three vessels. The mapping errors remain within a reasonable range ($1.72\pm0.62$), demonstrating that the proposed system achieves good alignment between US and CBCT.
\revision{The achieved precision is sufficient for liver ablation procedures, as study indicated that a needle placement accuracy below 5 mm ensures effective ablation while minimizing damage to healthy tissue ~\cite{nicolau2009augmented}.
These errors arise from multiple sources, including calibration errors between the camera and robot coordinate systems, the camera and CBCT coordinate systems, and the US and robot arm coordinate systems. 
As reported in \cite{sun2018robot}, using a similar hardware setup and the same calibration method, the error was measured at $0.573$ mm. 
Since the current mapping between CBCT and US is solely based on calibration processes and does not account for the alignment of anatomical features across modalities, future work could further minimize the mapping error by incorporating feature-based registration methods.}

\subsection{User study}\label{subsec3}

\revision{
Five volunteers participated in the experiment. Three were familiar with both needle insertion and robotic US systems, while the remaining two had experience with the US system but lacked familiarity with needle insertion and robotic systems. Prior to the experiment, all participants underwent training to familiarize themselves with the proposed system.
}

To compare the conventional US-guided needle insertion workflow with the workflow guided by the proposed CBCT-US fusion, two test scenarios are designed. In the first scenario, needle insertion is performed using only US guidance. The robotic arm is set to hand-guide mode, allowing users to manually manipulate the US probe into the correct position. The objective is for the needle, inserted through the needle holder, to successfully reach the lesion center while avoiding nearby vessels. Once the optimal US imaging plane is identified, the robotic arm is fixed in place, so that the users can free both hands for the needle insertion procedure. In the second scenario, the needle insertion is guided by the robotic CBCT-US system, where the localization process is devided into out-of-plane and in-plane localization as described in Sec.~\ref{sec:needle_planning}. The user is asked to determine the proper CT slice that passes through the lesion center while avoiding the occlusion of ribs. On the resulting US images, two points are manually selected by the users to allocate the needle insertion trajectory. The two approaches are compared in terms of time efficiency, accuracy, and success rates.

\begin{table}[ht!]
\centering
\caption{User Study Statistics: Comparison of Freehand and Robotic CBCT-US Methods}\label{tab:user_study}
\begin{tabular}{c|ccc}
\toprule
                &Searching time (s) &Lesion center deviation (mm) &Success rate \\
\midrule
Freehand        &57.11 $\pm$ 30.28 &7.89 $\pm$ 4.09  &65\%  \\
Robotic CBCT-US  &23.32 $\pm$ 5.90 &2.89 $\pm$ 1.68  &95\%  \\
\bottomrule
\end{tabular}
\end{table}

\par
The results are shown in Tab.~\ref{tab:user_study}. The searching time is defined as the time the volunteers spent locating the appropriate US imaging plane for needle insertion. Since the needle itself is not tracked, the deviation from the lesion center is estimated by calculating the distance between the expected needle trajectory and the lesion center. The expected needle trajectory is defined as the ideal path the needle would follow when precisely guided by the needle holder, which is rigidly mounted beside the US probe. To determine the success rate, an insertion attempt is considered unsuccessful if it involved contact with vessels or ribs, or if more than two retrials were necessary to achieve a proper insertion. With the guidance from the proposed robotic CBCT-US system, the searching time is reduced by more than $50\%$ compared to the free-hand US guidance. Additionally, the deviation from the lesion center is significantly improved by 5 mm with the assistance of the proposed system. The higher success rate further highlights the superiority of the CBCT-US fusion system, demonstrating its enhanced precision and efficiency in needle insertion tasks.

\subsection{Registration Accuracy After Repositioning}\label{subsec4}

In this section we evaluate the accuracy of the registation after repositioning of the CBCT device without system re-calibration. Based on the tracking information provided by the robotic CBCT about its movements, the transformation between the original CBCT coordinate system and the new CBCT coordinate system ($T^{c_{old}}_{c_{new}}$) can be determined. Then the calibration matrix after repositioning ($T^{u}_{c_{new}}$) can be updated as:
$T^{u}_{c_{new}}  =  T^{u}_{b} \, T^{b}_{o} \, T^{o}_{c_{old}} \, T^{c_{old}}_{c_{new}}$. The ground truth calibration is determined by re-performing the calibration process, as described in Sec.~\ref{sec:calibration} at the new position of the CBCT device. The system was evaluated by moving the LoopX device to five random positions from the initial position in two orthogonal directions in $\pm30~mm$ and $\pm10~mm$, respectively. The registration error after repositioning is $2.59\pm0.76$ mm and $0.75^{\circ}\pm0.42^{\circ}$ on average in translation and rotation, respectively. These results indicate that the system maintains a reasonable level of accuracy without requiring re-calibration after repositioning.  


\section{\revision{Discussion}}

\par
\revision{
Our current approach provides a practical solution for achieving registration initialization between the two modalities, with needle insertion showcasing the system’s potential for clinical application. The primary aim has been to demonstrate its feasibility in this scenario, while the needle insertion task itself is not our main focus. Compared to single-modality robotic needle insertion methods, our system combines the real-time soft tissue detail of US with the 3D anatomical context of CBCT, offering the potential for improved visualization.}
\par
\revision{Currently, the system operates in a semi-automated mode, where the robotic US system positions itself based on fused CBCT and US modalities to guide manual needle insertion through a needle holder. In the future, the system could be fully automated by incorporating a robotic needle insertion mechanism, which may improve accuracy by ensuring consistent and precise movements, compared to the variability introduced by human manipulation. The semi-automated mode serves as a proof of concept, laying the groundwork for advancements in fully automated approaches to enhance clinical outcomes.}
\par
\revision{Moving forward, we will prioritize integrating the image-based registration~\cite{jiang2024class} to enhance both accuracy and practicality. Additionally, addressing and compensating for registration errors caused by internal anatomy motion~\cite{jiang2022precise} and deformation~\cite{jiang2023defcor} due to breathing will be crucial for adapting the system to more complex and realistic scenarios. To facilitate adoption in clinical settings, we will evaluate the system’s safety and usability and seek feedback from both surgeons and patients. Future directions could also include exploring further automation of surgical tasks and integrating artificial intelligence to optimize procedural decision-making~\cite{bi2022vesnet, jiang2024intelligent}, ultimately making the system more intelligent and reliable.}


\section{Conclusion}

In this work, we presented a robotic dual-modality imaging system that integrates robotic CBCT and robotic US to provide enhanced fusion guidance for clinical procedures. The system allows for pre-calibrated and dynamically co-registered CBCT and US images, enabling registration-free, multi-modality image fusion. By leveraging both modalities, our system combines CBCT with the real-time soft tissue visualization and Doppler flow information from US. 


The effectiveness of this approach was validated in a needle insertion scenario simulating complex anatomical structures. The results demonstrate that the fused imaging system is able to enhance the accuracy and safety of needle insertion procedures. Our user study \revision{indicated an} improvement in time efficiency, lesion targeting accuracy, and overall success rate compared to traditional freehand US-guided workflow. Moreover, the system is able to maintain registration accuracy even after repositioning the CBCT device, showing the capability of co-registration for mobile imaging systems. 

In conclusion, our proposed robotic CBCT-US system paves the way for advanced and automated interventions in a clinical setting, with potential beyond just needle insertions. \revision{The integration of two robotic systems provides a promising platform for various clinical applications. }
This dual-modality system could be applied to more complex interventions, where precise navigation and multi-modal imaging are essential. Morworeover, the system’s ability to maintain registration accuracy after repositioning opens the door for more flexible and adaptive intra-operative imaging workflows. 

\section*{Declarations}
\subsection*{Funding} 
Partial financial support was received from Brainlab Vaskuläre Chirurgie.

\subsection*{Conflict of interest} 
The authors declare no conflict of interest.

\subsection*{Informed consent} 
Informed consent was obtained from all individual participants included in the study.



\bibliography{sn-bibliography}


\begin{thebibliography}{27}
\ifx \bisbn   \undefined \def \bisbn  #1{ISBN #1}\fi
\ifx \binits  \undefined \def \binits#1{#1}\fi
\ifx \bauthor  \undefined \def \bauthor#1{#1}\fi
\ifx \batitle  \undefined \def \batitle#1{#1}\fi
\ifx \bjtitle  \undefined \def \bjtitle#1{#1}\fi
\ifx \bvolume  \undefined \def \bvolume#1{\textbf{#1}}\fi
\ifx \byear  \undefined \def \byear#1{#1}\fi
\ifx \bissue  \undefined \def \bissue#1{#1}\fi
\ifx \bfpage  \undefined \def \bfpage#1{#1}\fi
\ifx \blpage  \undefined \def \blpage #1{#1}\fi
\ifx \burl  \undefined \def \burl#1{\textsf{#1}}\fi
\ifx \doiurl  \undefined \def \doiurl#1{\url{https://doi.org/#1}}\fi
\ifx \betal  \undefined \def \betal{\textit{et al.}}\fi
\ifx \binstitute  \undefined \def \binstitute#1{#1}\fi
\ifx \binstitutionaled  \undefined \def \binstitutionaled#1{#1}\fi
\ifx \bctitle  \undefined \def \bctitle#1{#1}\fi
\ifx \beditor  \undefined \def \beditor#1{#1}\fi
\ifx \bpublisher  \undefined \def \bpublisher#1{#1}\fi
\ifx \bbtitle  \undefined \def \bbtitle#1{#1}\fi
\ifx \bedition  \undefined \def \bedition#1{#1}\fi
\ifx \bseriesno  \undefined \def \bseriesno#1{#1}\fi
\ifx \blocation  \undefined \def \blocation#1{#1}\fi
\ifx \bsertitle  \undefined \def \bsertitle#1{#1}\fi
\ifx \bsnm \undefined \def \bsnm#1{#1}\fi
\ifx \bsuffix \undefined \def \bsuffix#1{#1}\fi
\ifx \bparticle \undefined \def \bparticle#1{#1}\fi
\ifx \barticle \undefined \def \barticle#1{#1}\fi
\bibcommenthead
\ifx \bconfdate \undefined \def \bconfdate #1{#1}\fi
\ifx \botherref \undefined \def \botherref #1{#1}\fi
\ifx \url \undefined \def \url#1{\textsf{#1}}\fi
\ifx \bchapter \undefined \def \bchapter#1{#1}\fi
\ifx \bbook \undefined \def \bbook#1{#1}\fi
\ifx \bcomment \undefined \def \bcomment#1{#1}\fi
\ifx \oauthor \undefined \def \oauthor#1{#1}\fi
\ifx \citeauthoryear \undefined \def \citeauthoryear#1{#1}\fi
\ifx \endbibitem  \undefined \def \endbibitem {}\fi
\ifx \bconflocation  \undefined \def \bconflocation#1{#1}\fi
\ifx \arxivurl  \undefined \def \arxivurl#1{\textsf{#1}}\fi
\csname PreBibitemsHook\endcsname

\bibitem[\protect\citeauthoryear{Gardiazabal et~al.}{2014}]{gardiazabal2014towards}
\begin{bchapter}
\bauthor{\bsnm{Gardiazabal}, \binits{J.}},
\bauthor{\bsnm{Esposito}, \binits{M.}},
\bauthor{\bsnm{Matthies}, \binits{P.}},
\bauthor{\bsnm{Okur}, \binits{A.}},
\bauthor{\bsnm{Vogel}, \binits{J.}},
\bauthor{\bsnm{Kraft}, \binits{S.}},
\bauthor{\bsnm{Frisch}, \binits{B.}},
\bauthor{\bsnm{Lasser}, \binits{T.}},
\bauthor{\bsnm{Navab}, \binits{N.}}:
\bctitle{Towards personalized interventional spect-ct imaging}.
In: \bbtitle{Medical Image Computing and Computer-Assisted Intervention--MICCAI 2014: 17th International Conference, Boston, MA, USA, September 14-18, 2014, Proceedings, Part I 17},
pp. \bfpage{504}--\blpage{511}
(\byear{2014}).
\bcomment{Springer}
\end{bchapter}
\endbibitem

\bibitem[\protect\citeauthoryear{Monfardini et~al.}{2018}]{monfardini2018ultrasound}
\begin{barticle}
\bauthor{\bsnm{Monfardini}, \binits{L.}},
\bauthor{\bsnm{Orsi}, \binits{F.}},
\bauthor{\bsnm{Caserta}, \binits{R.}},
\bauthor{\bsnm{Sallemi}, \binits{C.}},
\bauthor{\bsnm{Della~Vigna}, \binits{P.}},
\bauthor{\bsnm{Bonomo}, \binits{G.}},
\bauthor{\bsnm{Varano}, \binits{G.}},
\bauthor{\bsnm{Solbiati}, \binits{L.}},
\bauthor{\bsnm{Mauri}, \binits{G.}}:
\batitle{Ultrasound and cone beam ct fusion for liver ablation}.
\bjtitle{International Journal of Hyperthermia}
\bvolume{35}(\bissue{1}),
\bfpage{500}--\blpage{504}
(\byear{2018})
\end{barticle}
\endbibitem

\bibitem[\protect\citeauthoryear{Monfardini et~al.}{2021}]{monfardini2021real}
\begin{barticle}
\bauthor{\bsnm{Monfardini}, \binits{L.}},
\bauthor{\bsnm{Gennaro}, \binits{N.}},
\bauthor{\bsnm{Orsi}, \binits{F.}},
\bauthor{\bsnm{Della~Vigna}, \binits{P.}},
\bauthor{\bsnm{Bonomo}, \binits{G.}},
\bauthor{\bsnm{Varano}, \binits{G.}},
\bauthor{\bsnm{Solbiati}, \binits{L.}},
\bauthor{\bsnm{Mauri}, \binits{G.}}:
\batitle{Real-time us/cone-beam ct fusion imaging for percutaneous ablation of small renal tumours: a technical note}.
\bjtitle{European Radiology}
\bvolume{31},
\bfpage{7523}--\blpage{7528}
(\byear{2021})
\end{barticle}
\endbibitem

\bibitem[\protect\citeauthoryear{Mason et~al.}{2019}]{mason2019combined}
\begin{barticle}
\bauthor{\bsnm{Mason}, \binits{S.A.}},
\bauthor{\bsnm{White}, \binits{I.M.}},
\bauthor{\bsnm{O'Shea}, \binits{T.}},
\bauthor{\bsnm{McNair}, \binits{H.A.}},
\bauthor{\bsnm{Alexander}, \binits{S.}},
\bauthor{\bsnm{Kalaitzaki}, \binits{E.}},
\bauthor{\bsnm{Bamber}, \binits{J.C.}},
\bauthor{\bsnm{Harris}, \binits{E.J.}},
\bauthor{\bsnm{Lalondrelle}, \binits{S.}}:
\batitle{Combined ultrasound and cone beam ct improves target segmentation for image guided radiation therapy in uterine cervix cancer}.
\bjtitle{International Journal of Radiation Oncology* Biology* Physics}
\bvolume{104}(\bissue{3}),
\bfpage{685}--\blpage{693}
(\byear{2019})
\end{barticle}
\endbibitem

\bibitem[\protect\citeauthoryear{Kim et~al.}{2019}]{kim2019design}
\begin{barticle}
\bauthor{\bsnm{Kim}, \binits{M.D.}},
\bauthor{\bsnm{Lee}, \binits{K.-H.}},
\bauthor{\bsnm{Lee}, \binits{S.-M.}},
\bauthor{\bsnm{Koo}, \binits{J.C.}},
\bauthor{\bsnm{Ji}, \binits{S.-H.}}:
\batitle{Design and development of a mobile robotic ct system for intraoperative use}.
\bjtitle{IEEE/ASME Transactions on Mechatronics}
\bvolume{24}(\bissue{1}),
\bfpage{395}--\blpage{405}
(\byear{2019})
\end{barticle}
\endbibitem

\bibitem[\protect\citeauthoryear{Weir et~al.}{2015}]{weir2015dosimetric}
\begin{barticle}
\bauthor{\bsnm{Weir}, \binits{V.J.}},
\bauthor{\bsnm{Zhang}, \binits{J.}},
\bauthor{\bsnm{Bruner}, \binits{A.P.}}:
\batitle{Dosimetric characterization and image quality evaluation of the airo mobile ct scanner}.
\bjtitle{Journal of X-ray science and technology}
\bvolume{23}(\bissue{3}),
\bfpage{373}--\blpage{381}
(\byear{2015})
\end{barticle}
\endbibitem

\bibitem[\protect\citeauthoryear{Ebinger et~al.}{2015}]{ebinger2015mobile}
\begin{barticle}
\bauthor{\bsnm{Ebinger}, \binits{M.}},
\bauthor{\bsnm{Fiebach}, \binits{J.B.}},
\bauthor{\bsnm{Audebert}, \binits{H.J.}}:
\batitle{Mobile computed tomography: prehospital diagnosis and treatment of stroke}.
\bjtitle{Current opinion in neurology}
\bvolume{28}(\bissue{1}),
\bfpage{4}--\blpage{9}
(\byear{2015})
\end{barticle}
\endbibitem

\bibitem[\protect\citeauthoryear{Tsang et~al.}{2015}]{tsang2015real}
\begin{barticle}
\bauthor{\bsnm{Tsang}, \binits{R.K.}},
\bauthor{\bsnm{Sorger}, \binits{J.M.}},
\bauthor{\bsnm{Azizian}, \binits{M.}},
\bauthor{\bsnm{Holsinger}, \binits{C.F.}}:
\batitle{Real-time navigation in transoral robotic nasopharyngectomy utilizing on table fluoroscopy and image overlay software: a cadaveric feasibility study}.
\bjtitle{Journal of robotic surgery}
\bvolume{9},
\bfpage{311}--\blpage{314}
(\byear{2015})
\end{barticle}
\endbibitem

\bibitem[\protect\citeauthoryear{Tanaka et~al.}{2024}]{tanaka2024low}
\begin{barticle}
\bauthor{\bsnm{Tanaka}, \binits{M.}},
\bauthor{\bsnm{Schol}, \binits{J.}},
\bauthor{\bsnm{Sakai}, \binits{D.}},
\bauthor{\bsnm{Sako}, \binits{K.}},
\bauthor{\bsnm{Yamamoto}, \binits{K.}},
\bauthor{\bsnm{Yanagi}, \binits{K.}},
\bauthor{\bsnm{Hiyama}, \binits{A.}},
\bauthor{\bsnm{Katoh}, \binits{H.}},
\bauthor{\bsnm{Sato}, \binits{M.}},
\bauthor{\bsnm{Watanabe}, \binits{M.}}:
\batitle{Low radiation protocol for intraoperative robotic c-arm can enhance adolescent idiopathic scoliosis deformity correction accuracy and safety}.
\bjtitle{Global Spine Journal}
\bvolume{14}(\bissue{5}),
\bfpage{1504}--\blpage{1514}
(\byear{2024})
\end{barticle}
\endbibitem

\bibitem[\protect\citeauthoryear{Karius et~al.}{2024}]{karius2024first}
\begin{botherref}
\oauthor{\bsnm{Karius}, \binits{A.}},
\oauthor{\bsnm{Leifeld}, \binits{L.M.}},
\oauthor{\bsnm{Strnad}, \binits{V.}},
\oauthor{\bsnm{Fietkau}, \binits{R.}},
\oauthor{\bsnm{Bert}, \binits{C.}}:
First implementation of an innovative infra-red camera system integrated into a mobile cbct scanner for applicator tracking in brachytherapy—initial performance characterization.
Journal of Applied Clinical Medical Physics,
14364
(2024)
\end{botherref}
\endbibitem

\bibitem[\protect\citeauthoryear{Li et~al.}{2024}]{li2024invisible}
\begin{botherref}
\oauthor{\bsnm{Li}, \binits{C.}},
\oauthor{\bsnm{Huang}, \binits{D.}},
\oauthor{\bsnm{Karlas}, \binits{A.}},
\oauthor{\bsnm{Navab}, \binits{N.}},
\oauthor{\bsnm{Jiang}, \binits{Z.}}:
Invisible needle detection in ultrasound: Leveraging mechanism-induced vibration.
arXiv preprint arXiv:2403.14523
(2024)
\end{botherref}
\endbibitem

\bibitem[\protect\citeauthoryear{Jiang et~al.}{2024}]{jiang2024needle}
\begin{botherref}
\oauthor{\bsnm{Jiang}, \binits{Z.}},
\oauthor{\bsnm{Li}, \binits{X.}},
\oauthor{\bsnm{Chu}, \binits{X.}},
\oauthor{\bsnm{Karlas}, \binits{A.}},
\oauthor{\bsnm{Bi}, \binits{Y.}},
\oauthor{\bsnm{Cheng}, \binits{Y.}},
\oauthor{\bsnm{Au}, \binits{K.S.}},
\oauthor{\bsnm{Navab}, \binits{N.}}:
Needle segmentation using gan: Restoring thin instrument visibility in robotic ultrasound.
IEEE Transactions on Instrumentation and Measurement
(2024)
\end{botherref}
\endbibitem

\bibitem[\protect\citeauthoryear{Jiang et~al.}{2023}]{jiang2023robotic}
\begin{botherref}
\oauthor{\bsnm{Jiang}, \binits{Z.}},
\oauthor{\bsnm{Salcudean}, \binits{S.E.}},
\oauthor{\bsnm{Navab}, \binits{N.}}:
Robotic ultrasound imaging: State-of-the-art and future perspectives.
Medical image analysis,
102878
(2023)
\end{botherref}
\endbibitem

\bibitem[\protect\citeauthoryear{Bi et~al.}{}]{bi2024machine}
\begin{botherref}
\oauthor{\bsnm{Bi}, \binits{Y.}},
\oauthor{\bsnm{Jiang}, \binits{Z.}},
\oauthor{\bsnm{Duelmer}, \binits{F.}},
\oauthor{\bsnm{Huang}, \binits{D.}},
\oauthor{\bsnm{Navab}, \binits{N.}}:
Machine learning in robotic ultrasound imaging: Challenges and perspectives.
Annual Review of Control, Robotics, and Autonomous Systems
\textbf{7}
\end{botherref}
\endbibitem

\bibitem[\protect\citeauthoryear{von Haxthausen et~al.}{2021}]{von2021medical}
\begin{barticle}
\bauthor{\bsnm{Haxthausen}, \binits{F.}},
\bauthor{\bsnm{B{\"o}ttger}, \binits{S.}},
\bauthor{\bsnm{Wulff}, \binits{D.}},
\bauthor{\bsnm{Hagenah}, \binits{J.}},
\bauthor{\bsnm{Garc{\'\i}a-V{\'a}zquez}, \binits{V.}},
\bauthor{\bsnm{Ipsen}, \binits{S.}}:
\batitle{Medical robotics for ultrasound imaging: current systems and future trends}.
\bjtitle{Current robotics reports}
\bvolume{2},
\bfpage{55}--\blpage{71}
(\byear{2021})
\end{barticle}
\endbibitem

\bibitem[\protect\citeauthoryear{Huang et~al.}{2024}]{huang2024robot_qin}
\begin{botherref}
\oauthor{\bsnm{Huang}, \binits{Q.}},
\oauthor{\bsnm{Gao}, \binits{B.}},
\oauthor{\bsnm{Wang}, \binits{M.}}:
Robot-assisted autonomous ultrasound imaging for carotid artery.
IEEE Transactions on Instrumentation and Measurement
(2024)
\end{botherref}
\endbibitem

\bibitem[\protect\citeauthoryear{Li et~al.}{}]{li12autonomous}
\begin{botherref}
\oauthor{\bsnm{Li}, \binits{M.-D.}},
\oauthor{\bsnm{Lin}, \binits{X.-X.}},
\oauthor{\bsnm{Ruan}, \binits{S.M.}},
\oauthor{\bsnm{Ke}, \binits{W.-P.}},
\oauthor{\bsnm{Zhang}, \binits{H.-R.}},
\oauthor{\bsnm{Huang}, \binits{H.}},
\oauthor{\bsnm{Wu}, \binits{S.-H.}},
\oauthor{\bsnm{Cheng}, \binits{M.-Q.}},
\oauthor{\bsnm{Tong}, \binits{W.-J.}},
\oauthor{\bsnm{Hu}, \binits{H.-T.}}, et al.:
Autonomous robotic ultrasound scanning system: A key to enhancing image analysis reproducibility and observer consistency in ultrasound imaging.
Frontiers in Robotics and AI
\textbf{12},
1527686
\end{botherref}
\endbibitem

\bibitem[\protect\citeauthoryear{Jiang et~al.}{2021}]{jiang2021autonomous}
\begin{barticle}
\bauthor{\bsnm{Jiang}, \binits{Z.}},
\bauthor{\bsnm{Li}, \binits{Z.}},
\bauthor{\bsnm{Grimm}, \binits{M.}},
\bauthor{\bsnm{Zhou}, \binits{M.}},
\bauthor{\bsnm{Esposito}, \binits{M.}},
\bauthor{\bsnm{Wein}, \binits{W.}},
\bauthor{\bsnm{Stechele}, \binits{W.}},
\bauthor{\bsnm{Wendler}, \binits{T.}},
\bauthor{\bsnm{Navab}, \binits{N.}}:
\batitle{Autonomous robotic screening of tubular structures based only on real-time ultrasound imaging feedback}.
\bjtitle{IEEE Transactions on Industrial Electronics}
\bvolume{69}(\bissue{7}),
\bfpage{7064}--\blpage{7075}
(\byear{2021})
\end{barticle}
\endbibitem

\bibitem[\protect\citeauthoryear{Tsai et~al.}{1989}]{tsai1989new}
\begin{barticle}
\bauthor{\bsnm{Tsai}, \binits{R.Y.}},
\bauthor{\bsnm{Lenz}, \binits{R.K.}}, \betal:
\batitle{A new technique for fully autonomous and efficient 3 d robotics hand/eye calibration}.
\bjtitle{IEEE Transactions on robotics and automation}
\bvolume{5}(\bissue{3}),
\bfpage{345}--\blpage{358}
(\byear{1989})
\end{barticle}
\endbibitem

\bibitem[\protect\citeauthoryear{Jiang et~al.}{2023}]{jiang2023dopus}
\begin{botherref}
\oauthor{\bsnm{Jiang}, \binits{Z.}},
\oauthor{\bsnm{Duelmer}, \binits{F.}},
\oauthor{\bsnm{Navab}, \binits{N.}}:
Dopus-net: Quality-aware robotic ultrasound imaging based on doppler signal.
IEEE Transactions on Automation Science and Engineering
(2023)
\end{botherref}
\endbibitem

\bibitem[\protect\citeauthoryear{Nicolau et~al.}{2009}]{nicolau2009augmented}
\begin{barticle}
\bauthor{\bsnm{Nicolau}, \binits{S.}},
\bauthor{\bsnm{Pennec}, \binits{X.}},
\bauthor{\bsnm{Soler}, \binits{L.}},
\bauthor{\bsnm{Buy}, \binits{X.}},
\bauthor{\bsnm{Gangi}, \binits{A.}},
\bauthor{\bsnm{Ayache}, \binits{N.}},
\bauthor{\bsnm{Marescaux}, \binits{J.}}:
\batitle{An augmented reality system for liver thermal ablation: design and evaluation on clinical cases}.
\bjtitle{Medical image analysis}
\bvolume{13}(\bissue{3}),
\bfpage{494}--\blpage{506}
(\byear{2009})
\end{barticle}
\endbibitem

\bibitem[\protect\citeauthoryear{Sun et~al.}{2018}]{sun2018robot}
\begin{barticle}
\bauthor{\bsnm{Sun}, \binits{Y.}},
\bauthor{\bsnm{Jiang}, \binits{Z.}},
\bauthor{\bsnm{Qi}, \binits{X.}},
\bauthor{\bsnm{Hu}, \binits{Y.}},
\bauthor{\bsnm{Li}, \binits{B.}},
\bauthor{\bsnm{Zhang}, \binits{J.}}:
\batitle{Robot-assisted decompressive laminectomy planning based on 3d medical image}.
\bjtitle{IEEE Access}
\bvolume{6},
\bfpage{22557}--\blpage{22569}
(\byear{2018})
\end{barticle}
\endbibitem

\bibitem[\protect\citeauthoryear{Jiang et~al.}{2024}]{jiang2024class}
\begin{botherref}
\oauthor{\bsnm{Jiang}, \binits{Z.}},
\oauthor{\bsnm{Kang}, \binits{Y.}},
\oauthor{\bsnm{Bi}, \binits{Y.}},
\oauthor{\bsnm{Li}, \binits{X.}},
\oauthor{\bsnm{Li}, \binits{C.}},
\oauthor{\bsnm{Navab}, \binits{N.}}:
Class-aware cartilage segmentation for autonomous us-ct registration in robotic intercostal ultrasound imaging.
IEEE Transactions on Automation Science and Engineering
(2024)
\end{botherref}
\endbibitem

\bibitem[\protect\citeauthoryear{Jiang et~al.}{2022}]{jiang2022precise}
\begin{barticle}
\bauthor{\bsnm{Jiang}, \binits{Z.}},
\bauthor{\bsnm{Danis}, \binits{N.}},
\bauthor{\bsnm{Bi}, \binits{Y.}},
\bauthor{\bsnm{Zhou}, \binits{M.}},
\bauthor{\bsnm{Kroenke}, \binits{M.}},
\bauthor{\bsnm{Wendler}, \binits{T.}},
\bauthor{\bsnm{Navab}, \binits{N.}}:
\batitle{Precise repositioning of robotic ultrasound: Improving registration-based motion compensation using ultrasound confidence optimization}.
\bjtitle{IEEE Transactions on Instrumentation and Measurement}
\bvolume{71},
\bfpage{1}--\blpage{11}
(\byear{2022})
\end{barticle}
\endbibitem

\bibitem[\protect\citeauthoryear{Jiang et~al.}{2023}]{jiang2023defcor}
\begin{barticle}
\bauthor{\bsnm{Jiang}, \binits{Z.}},
\bauthor{\bsnm{Zhou}, \binits{Y.}},
\bauthor{\bsnm{Cao}, \binits{D.}},
\bauthor{\bsnm{Navab}, \binits{N.}}:
\batitle{Defcor-net: physics-aware ultrasound deformation correction}.
\bjtitle{Medical Image Analysis}
\bvolume{90},
\bfpage{102923}
(\byear{2023})
\end{barticle}
\endbibitem

\bibitem[\protect\citeauthoryear{Bi et~al.}{2022}]{bi2022vesnet}
\begin{barticle}
\bauthor{\bsnm{Bi}, \binits{Y.}},
\bauthor{\bsnm{Jiang}, \binits{Z.}},
\bauthor{\bsnm{Gao}, \binits{Y.}},
\bauthor{\bsnm{Wendler}, \binits{T.}},
\bauthor{\bsnm{Karlas}, \binits{A.}},
\bauthor{\bsnm{Navab}, \binits{N.}}:
\batitle{Vesnet-rl: Simulation-based reinforcement learning for real-world us probe navigation}.
\bjtitle{IEEE Robotics and Automation Letters}
\bvolume{7}(\bissue{3}),
\bfpage{6638}--\blpage{6645}
(\byear{2022})
\end{barticle}
\endbibitem

\bibitem[\protect\citeauthoryear{Jiang et~al.}{2024}]{jiang2024intelligent}
\begin{barticle}
\bauthor{\bsnm{Jiang}, \binits{Z.}},
\bauthor{\bsnm{Bi}, \binits{Y.}},
\bauthor{\bsnm{Zhou}, \binits{M.}},
\bauthor{\bsnm{Hu}, \binits{Y.}},
\bauthor{\bsnm{Burke}, \binits{M.}},
\bauthor{\bsnm{Navab}, \binits{N.}}:
\batitle{Intelligent robotic sonographer: Mutual information-based disentangled reward learning from few demonstrations}.
\bjtitle{The International Journal of Robotics Research}
\bvolume{43}(\bissue{7}),
\bfpage{981}--\blpage{1002}
(\byear{2024})
\end{barticle}
\endbibitem

\end{thebibliography}

\end{document}